\documentclass{article}

\usepackage{tikz}
\usepackage{amsmath}
\usetikzlibrary{positioning,arrows.meta,calc,fit,backgrounds,shadows.blur}
\usepackage{subfig}
\usepackage{subcaption}

 \usepackage[dblblindworkshop, final]{neurips_2025}

\workshoptitle{Foundation Models for the Brain and Body}

\usepackage[utf8]{inputenc} 
\usepackage[T1]{fontenc}    
\usepackage{hyperref}       
\usepackage{url}            
\usepackage{booktabs}       
\usepackage{amsfonts}       
\usepackage{nicefrac}       
\usepackage{microtype}      
\usepackage{xcolor}         
\usepackage{graphicx} 

\title{Scalable Diffusion Transformer for Conditional 4D fMRI Synthesis}

\author{%
  Jungwoo Seo \\
  Seoul National University \\
  \texttt{jungwoo.seo95@gmail.com} \\
  \And
  David Keetae Park \\
  Brookhaven National Laboratory \\
  \texttt{dpark1@bnl.gov} \\
  \And
  Shinjae Yoo$^{*}$ \\
  Brookhaven National Laboratory \\
  \texttt{sjyoo@bnl.gov} \\
  \And
  Jiook Cha$^{*}$ \\
  Seoul National Laboratory \\
  \texttt{connectome@snu.ac.kr} \\
}

\begin{document}

\maketitle

\renewcommand{\thefootnote}{*}
\footnotetext{Co-corresponding author}
\renewcommand{\thefootnote}{\arabic{footnote}}

\begin{abstract}
  Generating whole-brain 4D fMRI sequences conditioned on cognitive tasks remains challenging due to the high-dimensional, heterogeneous BOLD dynamics across subjects/acquisitions and the lack of neuroscience-grounded validation. We introduce the first diffusion transformer for voxelwise 4D fMRI conditional generation, combining 3D VQ-GAN latent compression with a CNN–Transformer backbone and strong task conditioning via AdaLN-Zero and cross-attention. On HCP task fMRI, our model reproduces task-evoked activation maps, preserves the inter-task representational structure observed in real data (RSA), achieves perfect condition specificity, and aligns ROI time-courses with canonical hemodynamic responses. Performance improves predictably with scale, reaching task-evoked map correlation of 0.83 and RSA of 0.98, consistently surpassing a U-Net baseline on all metrics. By coupling latent diffusion with a scalable backbone and strong conditioning, this work establishes a practical path to conditional 4D fMRI synthesis, paving the way for future applications such as virtual experiments, cross-site harmonization, and principled augmentation for downstream neuroimaging models.
\end{abstract}

\section{Introduction}

Task-based fMRI (task-fMRI) offers a powerful lens on the spatio-temporal dynamics underlying cognition. A critical frontier in cognitive neuroscience is to develop generative models that capture the mapping between cognitive processes and whole-brain activity patterns. Such models would not only synthesize realistic fMRI data but also enable in-silico experiments, providing a means to probe brain dynamics under controlled cognitive manipulations~\cite{pinaya2022brain, jain2024computational}.

Despite this potential, generating voxel-level whole-brain 4D task-fMRI remains unsolved. Unlike static 3D structural MRI, where diffusion models have successfully synthesized high-fidelity brain anatomies~\cite{pinaya2022brain, khader2023denoising}, fMRI poses dual challenges: extreme dimensionality and substantial inter-subject and acquisition-related variability. These factors often obscure subtle task-evoked signals (Appendix~\ref{appx:subject_acquisition_variabilities}), as emphasized in prior studies~\cite{miller2009unique, gratton2018functional, mori2018effect, wang2022effect}. Consequently, earlier work has avoided voxel-level dynamics (see more related works on Appendix~\ref{app:related_work}), instead focusing on simplified representations such as region-of-interest (ROI) time series~\cite{hu2025synthesizing, koppe2019identifying}, functional connectivity~\cite{orlichenko2024demographic}, or static 3D activation maps~\cite{bao2025mindsimulator, mai2025synbrain}. To date, no method has successfully generated task-conditioned, whole-brain 4D fMRI data using modern generative architectures.

In this work, we introduce the first conditional diffusion transformer for voxel-wise, 4D task-fMRI synthesis. Our approach is designed to be both computationally efficient and scalable, directly addressing the challenges of modeling high-dimensional spatio-temporal brain dynamics. Our contributions are summarized as follows:
\begin{itemize}
\item \textbf{First Conditional 4D fMRI Generative Model}. We present the first voxel-level, whole-brain 4D diffusion model conditioned on cognitive tasks, enabling realistic synthesis of spatio-temporal brain dynamics.
\item \textbf{Scalable Hybrid Architecture}. We propose a latent diffusion transformer that combines 3D VQ-GAN compression, a CNN–Transformer hybrid backbone, and enhanced conditional injection via AdaLN and cross-attention. Ablation studies validate the role of each component.
\item \textbf{Neuroscience-aligned Evaluation}. We evaluate our model on seven HCP task-fMRI paradigms and introduce neuroscience-aligned metrics—brain activation map correlation, representational similarity analysis (RSA of inter-task structures), and condition specificity—to assess the fidelity of generated task-evoked brain responses. Our model consistently outperforms a UNet diffusion baseline.

\end{itemize}

\begin{figure}[t!]  
    \centering
    \includegraphics[width=1.0\linewidth]{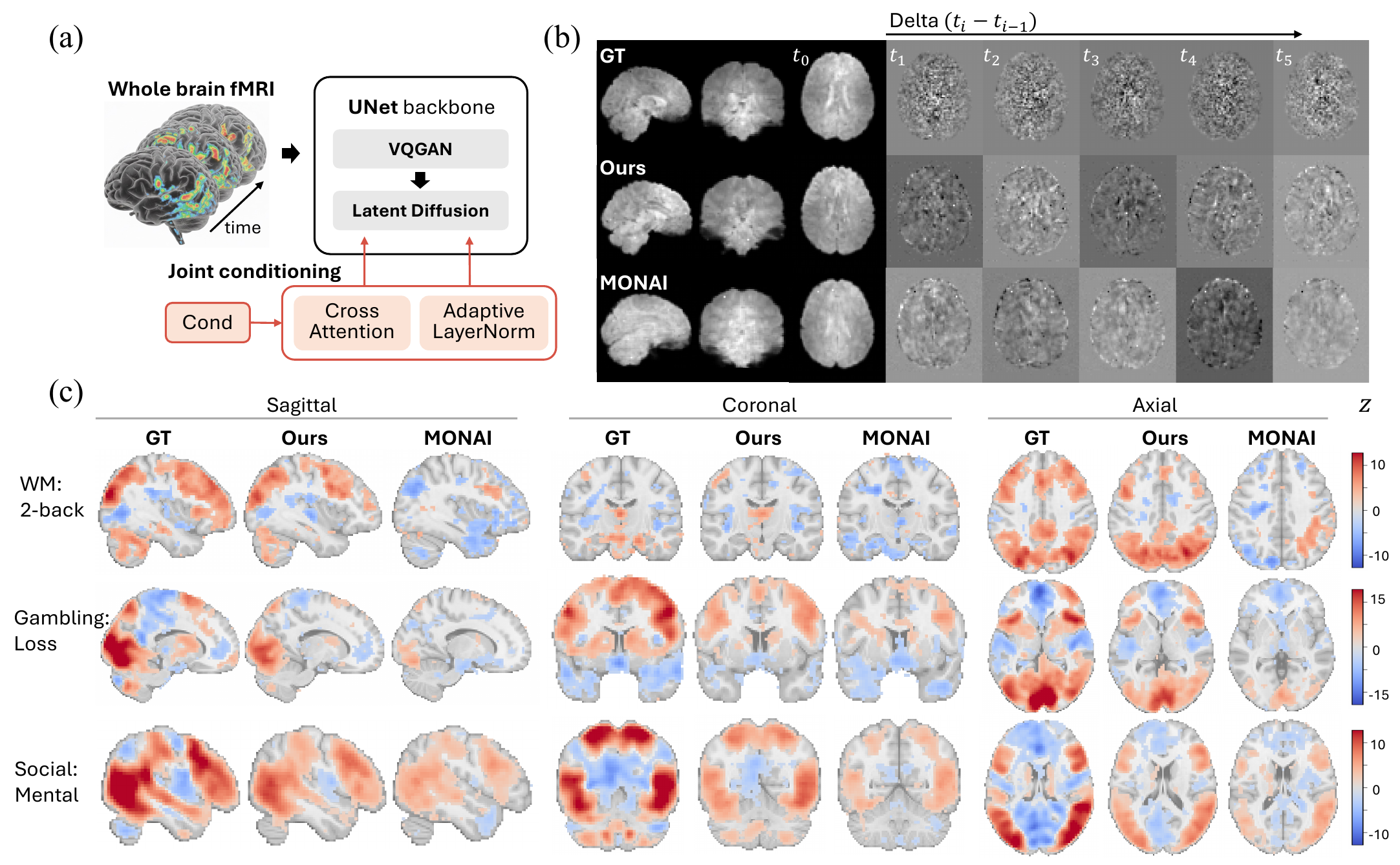}
    \caption{Overview and sampled visual results compared against MONAI~\cite{pinaya2023generative}. (a) Our architecture follows recent advances in VQGAN~\citep{kim2024adaptive} and Latent Diffusion~\citep{rombach2022high}, equipped with joint conditioning for strong conditional generative performances. (b) Compared to the the closest baseline, MONAI, our model generates superior spatial details and (c) \emph{conditional} temporal BOLD dynamics as revealed by a group-level GLM activation  map (Section~\ref{sec:eval_metrics}).}
    \label{fig:generation}
\end{figure}
\vspace*{-0.4cm}

\section{Methods}
\paragraph{Problem Formulation.}
We consider the task of conditional generation of 4D task-fMRI volumes given a task label $c$. 
Each fMRI instance is a spatio-temporal tensor $x \in \mathbb{R}^{H \times W \times D \times T}$, 
where $H, W, D$ denote spatial dimensions and $T$ is the number of time points. 
Our goal is to model the conditional distribution $p_\theta(x \mid c)$, 
from which synthetic task-evoked fMRI volumes can be sampled to faithfully reproduce spatio-temporal brain dynamics under condition $c$.

\paragraph{Latent Diffusion Modeling}
We adopt a denoising diffusion probabilistic model (DDPM)~\cite{ho2020denoising}. A direct diffusion in voxel space is computationally infeasible due to the extremely high dimensionality of fMRI volumes and the limited availability of training data. Following a latent diffusion framework~\citep{rombach2022high}, each fMRI volume is first compressed by a pretrained 3D VQ-GAN~\citep{kim2024adaptive}, yielding 
$z \in \mathbb{R}^{C \times (H/4)\times(W/4)\times(D/4)\times T}$, where $C$ denotes the latent channel dimension. 

The forward process gradually adds Gaussian noise:  
$z_t = \sqrt{\bar{\alpha}_t}\, z_0 + \sqrt{1-\bar{\alpha}_t}\,\epsilon$, 
with $\epsilon \sim \mathcal{N}(0,I)$, 
$\bar{\alpha}_t=\prod_{s=1}^t(1-\beta_s)$, and schedule $\{\beta_t\}_{t=1}^T$.  
The reverse process is parameterized as  
\[
p_\theta(z_{t-1}\mid z_t,c)=\mathcal{N}\!\left(z_{t-1};\,\mu_\theta(z_t,t,c),\,\sigma_t^2 I\right),
\]
where the network predicts $\epsilon_\theta(z_t,t,c)$, from which $\mu_\theta$ is derived. 
Training minimizes mean-squared error between injected noise $\epsilon$ and predicted noise  $\epsilon_\theta(z_t,t,c)$ with the simple objective
$\mathcal{L}_{\text{simple}}=\mathbb{E}_{z_0,\epsilon,t,c}\!\left[\|\epsilon-\epsilon_\theta(z_t,t,c)\|^2\right]
$. 
Generation is performed by sampling $z_T\!\sim\!\mathcal{N}(0,I)$ and iteratively denoised to $z_0$, which is decoded by the VQ-GAN decoder into a 4D fMRI volume.

\paragraph{CNN--Transformer Hybrid Diffusion Backbone.}
To effectively model high-dimensional spatio-temporal fMRI under limited data, our design seeks a balance between computational efficiency, inductive bias, and scalability. 
We therefore adopt a hierarchical CNN--Transformer architecture (Appendix~\ref{appx:neural_architectures}), inspired by several previous designs \citep{nawrot2021hierarchical, hoogeboom2023simple, kim2023swiftswin4dfmri, crowson2024scalable}. Within earlier layers, convolutional residual blocks provide strong local spatio-temporal inductive bias, thereby reducing computation cost and enabling stable training with limited data. For later layers, transformer blocks with global attention capture long-range dependencies across space and time, leveraging the strong scalability of diffusion transformers \citep{peebles2023scalablediffusionmodelstransformers}.
These components are organized in a UNet-like hierarchy, where features from different resolutions are fused through concatenation, producing multi-scale representations that integrate local detail with global context—an essential property for modeling distributed brain activations in fMRI.

\paragraph{Conditioning Mechanisms.}
To mitigate inter-subject and acquisition variability and amplify subtle task-specific signals, we adopt two complementary conditioning mechanisms. First, adaptive normalization: Transformer blocks use \textit{AdaLN-Zero}\citep{peebles2023scalablediffusionmodelstransformers}, modulating LayerNorm scale and shift from the condition $c$, while convolutional residual blocks apply \textit{FiLM}\citep{perez2018film} for condition-dependent modulation. Second, cross-attention directly exchanges information between condition embeddings and latent tokens, injecting stronger task-specific signals that help overcome task-agnostic variability.
As shown in ablation studies (Table~\ref{tab:ablation-arch}), combining normalization-based and cross-attention conditioning yields more faithful task-specific synthesis than using AdaLN-Zero alone.

\section{Experiments}
\subsection{Experimental Setup}

\paragraph{Dataset and Implementation}
We evaluate our model on the Human Connectome Project (HCP) task-fMRI dataset~\citep{van2013wu, barch2013function}, which comprises seven paradigms: working memory, emotion, language, motor, relational, social, and gambling. For each paradigm, we select one representative condition that reliably elicits the intended cognitive process, defining seven task-condition labels (Appendix~\ref{appx:task_conditions}). We use minimally preprocessed fMRI data~\citep{glasser2013minimal} and apply further preprocessing to reduce computational burden: downsampling to $3 \times 3 \times 3$ mm$^3$ spatial resolution, resampling to TR = 1.44s, and cropping background voxels. Since HCP task fMRI follows a block design, we treat each condition block as an instance, extracting approximately 18s ($\sim$12 TRs) from onset for each instance. This yields 34,632 instances from 1,083 participants. Data are split by subject into 90/5/5 train/validation/test sets, resulting in 31,168 training instances from 975 subjects.

All models were trained with AdamW for 400k steps using a linear diffusion noise schedule, class dropout for classifier-free guidance, and exponential moving average (EMA) for sampling. Training used a single NVIDIA A100 (40GB) with bfloat16 mixed precision. Full hyperparameters and model specifications for both our models and MONAI baselines are provided in Appendix~\ref{appx:implementation_details}.

\paragraph{Baseline Model.} To benchmark our model, we selected a diffusion-based approach, which represents the state-of-the-art in generative modeling. Specifically, we employed a 3D U-Net conditional diffusion model from the MONAI generative package \citep{pinaya2023generative}, a framework well-validated for structural MRI synthesis (e.g., \citep{pinaya2022brain, puglisi2024enhancing}). To ensure a direct and fair comparison, we adapted this 3D baseline for 4D fMRI generation using the exact same strategy as our proposed method: stacking temporal frames along the channel dimension and operating within the identical latent space from our VQ-GAN encoder. We note that while a prior study used an $\alpha$-GAN for fMRI generation \citep{wang2023learning}, its implementation is not publicly available, preventing its replication for our analysis.

\paragraph{Scalability Study.}
We investigated scalability by progressively increasing model capacity from tens to hundreds of millions of trainable parameters, by varying three factors: model width, condition embedding dimension, and number of attention heads (Appendix~\ref{appx:implementation_details}). 
This allows us to examine whether generative performance scales predictably with model capacity, a hallmark of foundation models.

\subsection{Neuroscience-aligned Evaluation Metrics}
\label{sec:eval_metrics}

Standard image metrics such as FID or Inception Score assess low-level realism but fail to test whether generated fMRI preserves task-specific spatio-temporal dynamics. We therefore introduce three neuroscience-aligned metrics: (i) \textbf{brain activation map correlation (Corr):} voxelwise Pearson correlation between real and synthetic group-level generalized linear model (GLM) contrast maps (\emph{z}-maps), assessing fidelity of task-evoked activations. (ii) \textbf{representational similarity analysis (RSA):} correlation between the off-diagonal entries of real and synthetic representational dissimilarity matrices (RDMs), measuring preservation of inter-task representational geometry. (iii) \textbf{condition specificity (Top-1 Accuracy):} fraction of generated samples whose activation map best matches the correct real task map, with 1 indicating perfect specificity.

Together, these metrics evaluate fidelity at three levels: voxelwise activations, inter-task structure, and task-specific identifiability. Full formulations are provided in Appendix~\ref{appx:eval_details}.

\subsection{Results}

\begin{figure}[h]  
    \centering
    \includegraphics[width=1.0\linewidth]{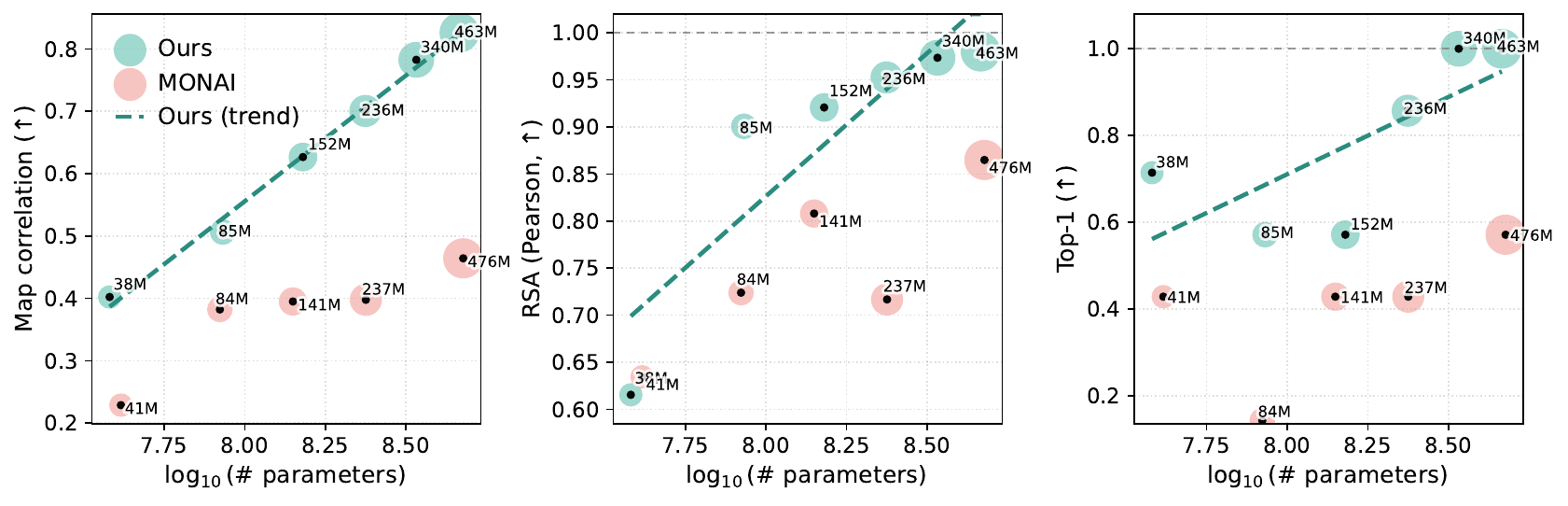}
    \caption{Generative performance as a function of model capacity.
Performance is evaluated using three neuroscience-aligned metrics: (a) GLM activation map correlation, (b) Representational similarity analysis (RSA, Pearson), and (c) Condition specificity (Top-1 Accuracy).}
    \label{fig:scaling}
\end{figure}
\vspace*{-0.2cm}

\paragraph{Qualitative Evaluations.}
Fig.~\ref{fig:generation} provides an overview of our model and representative qualitative results. 
Compared to the baseline (MONAI), our model generates fMRI sequences with sharper spatial detail and more realistic temporal BOLD dynamics. 
In particular, group-level GLM contrasts reveal that condition-specific task activations are faithfully reproduced across multiple paradigms (Fig.~\ref{fig:generation}c). 
These results demonstrate that the combination of latent diffusion with hybrid CNN--Transformer backbones and joint conditioning yields substantial improvements in generative fidelity and task-relevant modeling.

\paragraph{Model Scaling Behavior.}
Fig.~\ref{fig:scaling} summarizes how generative performance scales with model capacity. Across all three metrics, our conditional diffusion transformer exhibits consistent gains as the number of trainable parameters scales. GLM activation map correlation steadily improves, indicating that larger models more faithfully reproduce task-evoked spatiotemporal activation patterns. Similarly, RSA scores hits the theoretical maximum level of 1.0, showing that inter-task representational relationship is increasingly preserved. Condition specificity also rises with scale, reaching perfect Top-1 Accuracy for 340~M and larger models.

Importantly, these improvements follow a clear scaling trend, consistent with observations from foundation model research in vision and language. In contrast, the MONAI baseline, despite having comparable parameter count, lags behind our models in all metrics, underscoring the effectiveness of our hybrid diffusion-transformer design for modeling spatio-temporal brain dynamics.

\paragraph{Ablations.}

\begin{table}[h]
\centering
\caption{Ablation studies on backbone and conditioning mechanisms. For conditioning, the full and no cross-attention models differ in parameter count due to the presence of cross-attention, but model widths were matched for fair comparison.}
\label{tab:ablation-arch}
\begin{tabular}{lrrrr}
\toprule
\textbf{Model Backbones} & \multicolumn{1}{c}{Params} & \multicolumn{1}{c}{Corr (↑)} & \multicolumn{1}{c}{Top-1 (↑)} & \multicolumn{1}{c}{RSA (↑)} \\
\cmidrule(lr){1-1}\cmidrule(lr){2-5}
Hybrid (CNN early + Transformer mid/high) & 236.5M & \textbf{0.7006} & \textbf{0.8571} & \textbf{0.9526} \\
All-CNN                    & 235.0M & 0.6289          & 0.7143          & 0.9195 \\
All-Transformer           & 238.0M & 0.6734          & 0.7143          & 0.9448 \\
\cmidrule(lr){1-1}
\textbf{Conditioning Mechanisms}  \\
\cmidrule(lr){1-1}
Full conditioning (AdaLN-Zero + Cross-Attn) & 151.5M & \textbf{0.6267} & 0.5714 & \textbf{0.9207} \\
No cross-attention (AdaLN-Zero only)         &   
110.5M     & 0.5066          & \textbf{0.7143} & 0.9001 \\
\bottomrule
\end{tabular}
\end{table}

Table~\ref{tab:ablation-arch} details our architectural ablation study, which compares several backbone variants, and the conditioning mechanisms. An \emph{All-CNN} model yielded the weakest task fidelity, whereas an \emph{All-Transformer} model marginally improves performance. Our proposed hybrid design—employing convolutions in early stages and transformers in later stages—achieved the optimal balance between performance and efficiency. 

Beyond the backbone, removing cross-attention and relying solely on \emph{AdaLN-Zero} with \emph{FiLM}, despite reducing parameters, led to a notable performance drop. This result highlights the necessity of strong conditional injection to capture subtle task-specific activations. Such a mechanism is particularly critical for fMRI data, where task-evoked signals are often weak compared to inter-subject and acquisition variability. This suggests that future generative fMRI architectures will similarly benefit from explicit and powerful conditioning.
\section{Discussion and Conclusion}
We introduced a conditional diffusion transformer for whole-brain 4D fMRI synthesis. Our results highlight several insights. First, architectural design is critical: compared to U-Net baselines, transformer backbones provide strong scaling properties that enable synthesis of task-specific activations despite immense subject- and acquisition-level variability. Yet, a pure transformer backbone was suboptimal—best performance and efficiency arose from a hierarchical CNN-Transformer hybrid architecture, where convolutional residual blocks capture local structure in early stages and transformers integrate long-range dependencies in higher stages, offering both inductive bias and scalability under limited data. Second, strong task conditioning was found to be beneficial. Injecting conditioning signals via both AdaLN and cross-attention helped capture subtle condition-specific variance. 
Third, we observed that volume-wise compression with a 3D VQ-GAN, followed by latent diffusion, can also support effective 4D synthesis—producing spatially coherent and temporally plausible BOLD dynamics even without explicit 4D compression. 

Taken together, these findings suggest four principles for future 4D fMRI generative modeling: (1) ensure sufficient model capacity with scalable backbones, (2) introduce appropriate inductive bias when data are limited, (3) enforce strong conditional injection to capture task-specific signals, and (4) when high-quality 4D compression network is not available, volume-wise compression network such as 3D VQ-GAN \citep{kim2024adaptive} may serve as practical alternative.

Finally, diffusion transformers exhibit clear and predictable scaling laws (Fig.~\ref{fig:scaling}), consistent with trends seen in foundation models for vision and language. This indicates that generative neuroimaging may similarly benefit from scaling, paving the way toward fMRI generative foundation models. Such models could serve not only as data augmentation tools but also as scientific instruments—enabling virtual experiments, cross-site harmonization, simulation-based inference, and ultimately precision psychiatry. Future work will explore training across larger datasets, integration of multimodal signals, and downstream applications in neuroscience and medicine.

\section*{Acknowledgements}
This work was supported by the National Research Foundation of Korea (NRF) grant funded by the Korea government (MSIT) (No. 2021R1C1C1006503, RS-2023-00266787, RS-2023-00265406, RS-2024-00421268, RS-2024-00342301, RS-2024-00435727, NRF-2021M3E5D2A01022515, and NRF-2021S1A3A2A02090597), by the Creative-Pioneering Researchers Program through Seoul National University (No. 200-20240057, 200-20240135). Additional support was provided by the Institute of Information \& Communications Technology Planning \& Evaluation (IITP) grant funded by the Korea government (MSIT) [No. RS-2021-II211343, Artificial Intelligence Graduate School Program, Seoul National University] and by the Global Research Support Program in the Digital Field (RS-2024-00421268). This work was also supported by the Artificial Intelligence Industrial Convergence Cluster Development Project funded by the Ministry of Science and ICT and Gwangju Metropolitan City, by the Korea Brain Research Institute (KBRI) basic research program (25-BR-05-01), by the Korea Health Industry Development Institute (KHIDI) and the Ministry of Health and Welfare, Republic of Korea (HR22C1605), and by the Korea Basic Science Institute (National Research Facilities and Equipment Center) grant funded by the Ministry of Education (RS-2024-00435727). We acknowledge the National Supercomputing Center for providing supercomputing resources and technical support (KSC-2023-CRE-0568). An award for computer time was provided by the U.S. Department of Energy’s (DOE) ASCR Leadership Computing Challenge (ALCC). This research used resources of the National Energy Research Scientific Computing Center (NERSC), a DOE Office of Science User Facility, under ALCC award m4750-2024 and m4727, and supporting resources at the Argonne and Oak Ridge Leadership Computing Facilities, U.S. DOE Office of Science user facilities at Argonne National Laboratory and Oak Ridge National Laboratory. The Brookhaven National Laboratory team was supported by the U.S. Department of Energy (DOE), Office of Science (SC), Advanced Scientific Computing Research program under award DE-SC-0012704 and used resources of the National Energy Research Scientific Computing Center, a DOE Office of Science User Facility using NERSC award DDR-ERCAP0033558. We thank Seungjun Lee at Brookhaven National Laboratory for insightful comments that improved our study.

\bibliographystyle{unsrt}
\bibliography{neurips_2025}

\newpage
\appendix

\section{Appendix}

\subsection{Inter-subject and Acquisition-related Variabilities} \label{appx:subject_acquisition_variabilities}

\begin{figure}[h]  
    \centering
    \includegraphics[width=1.0\linewidth]{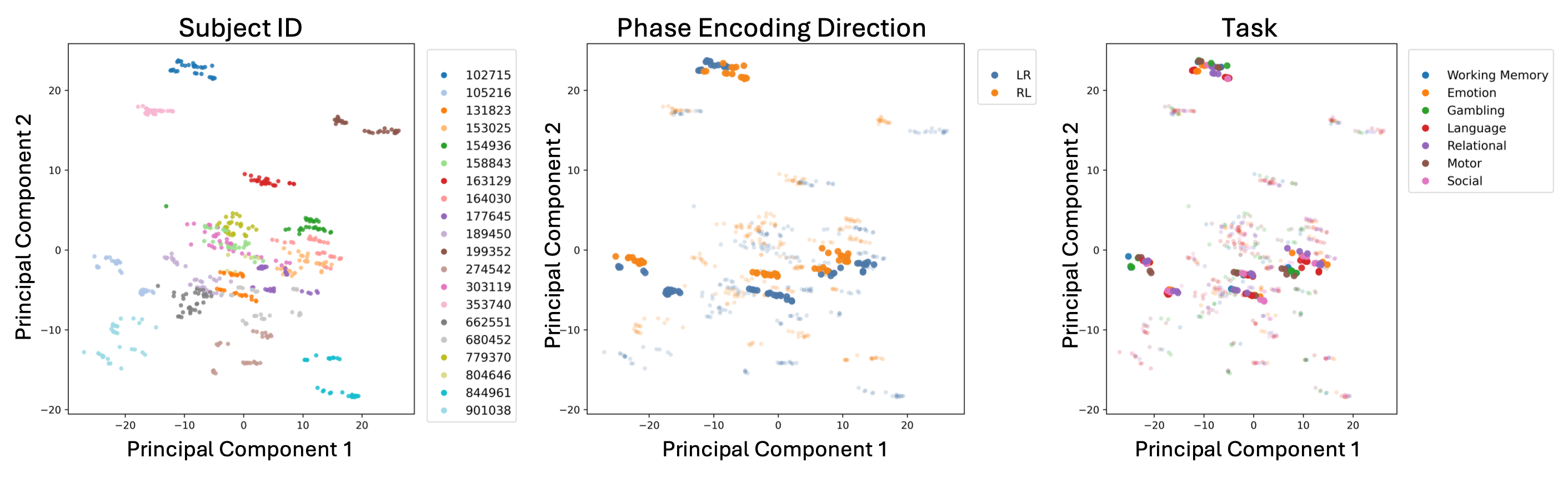}
    \caption{Principal component analysis (PCA) of HCP task-fMRI volumes.  
(Left) Clustering by subject shows that individual differences dominate the data variance.  
(Middle) Phase encoding direction (LR vs.~RL) explains the second-largest variability component.  
(Right) Task conditions contribute only weaker variance.  
These results indicate that subject- and acquisition-related factors obscure task-evoked signals, consistent with prior findings.}
    \label{fig:subject_acquisition_variabilities}
\end{figure}

To quantify the relative sources of variability in HCP task-fMRI data, we performed PCA on voxelwise BOLD signals restricted to the brain. 
As shown in Fig.~\ref{fig:subject_acquisition_variabilities}, the largest variance component is explained by \textbf{individual subject differences}, followed by \textbf{phase encoding direction} (LR vs.~RL), with \textbf{task-evoked variability} only emerging as a weaker source of variance. 
This analysis illustrates that inter-subject and acquisition-related factors dominate over task-related signals, underscoring the challenge of modeling condition-specific fMRI responses.

\subsection{Additional Related Work}
\label{app:related_work}

\paragraph{Generative Models for fMRI} Early approaches to fMRI synthesis relied on forward models with explicit statistical assumptions. Simulators like SimTB~\cite{erhardt2012simtb} and fmrisim~\cite{ellis2020facilitating} generate data from user-defined spatial sources and temporal dynamics. While useful for validating analysis pipelines, they cannot capture the complex, heterogeneous data distributions of real fMRI. More recent deep learning generative methods have learned data distributions from real data but have largely avoided the complexity of full 4D modeling. Many studies reduce the data to simplified representations, such as generating ROI- or parcel-based time series~\cite{hu2025synthesizing, koppe2019identifying, volkmann2024scalable}, functional connectivity matrices~\cite{orlichenko2024demographic}, or flattened spatial inputs~\cite{qiang2023functional}. Another popular approach is to generate static, 3D contrast maps derived from GLM analyses~\cite{bao2025mindsimulator, mai2025synbrain, zhuang2019fmri, tajini2021functional}. While successful in their respective scopes, these methods lose the rich, voxel-level spatio-temporal information that is critical for understanding brain dynamics. A single study attempted voxel-level 4D generation with an $\alpha$-GAN~\cite{rosca2017variational, wang2023learning}, but it was designed for data augmentation and, like many GANs, faced challenges in training stability and sample diversity.

\paragraph{Diffusion Models in Neuroimaging} Diffusion models~\cite{ho2020denoising} have emerged as the state-of-the-art for high-fidelity generative modeling. In neuroimaging, their success has been most prominent in the domain of structural MRI. By conditioning on covariates like age and sex, these models can synthesize realistic 3D brain anatomies~\cite{pinaya2022brain, khader2023denoising}. This has enabled a variety of applications, including data augmentation, modality conversion~\cite{kim2024adaptive}, super-resolution~\cite{wang2023inversesr}, and simulating disease progression~\cite{puglisi2024enhancing}. Inspired by their success in vision, Diffusion Transformers~\cite{peebles2023scalablediffusionmodelstransformers} have recently demonstrated remarkable performance and predictable scaling laws, establishing them as a promising architecture for foundation models. Their ability to handle long-range dependencies makes them particularly suitable for the spatio-temporal dynamics of 4D fMRI. Our work is the first to adapt and scale this architecture for the unique challenges of conditional, voxel-level fMRI generation.

\subsection{Neural Architectures} 
\label{appx:neural_architectures}
\begin{figure}[h]  
    \centering
    \includegraphics[width=0.7\linewidth]{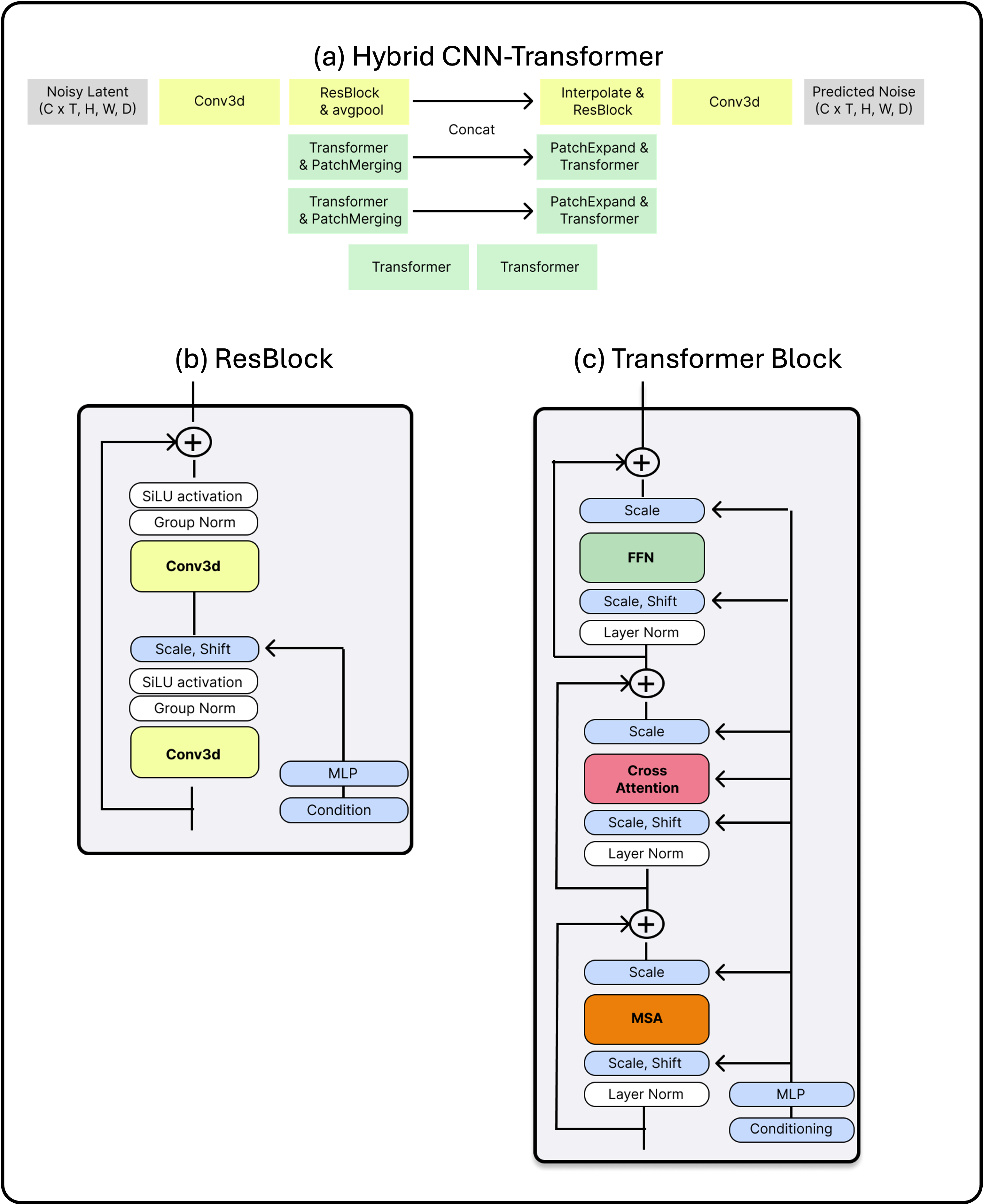}
    \caption{Detailed architecture of the proposed CNN–Transformer hybrid backbone. 
(a) UNet-style hierarchy integrating convolutional (yellow) and transformer (green) stages with downsampling/upsampling paths. 
(b) residual block with FiLM-based conditioning. 
(c) transformer block with AdaLN-Zero and cross-attention conditioning. This design balances between computational efficiency, inductive bias, and scalability.}
    \label{fig:architecture_appendix}
\end{figure}

Figure~\ref{fig:architecture_appendix} provides a schematic of our diffusion model backbone architecture. 

\clearpage
\subsection{Task-Condition Selection}
\label{appx:task_conditions}
The HCP task-fMRI dataset~\citep{van2013wu, barch2013function} consists of seven paradigms, each designed to elicit specific cognitive processes. 
For our experiments, we selected one representative condition per paradigm that reliably induces the intended cognitive process. 
Table~\ref{tab:task_conditions} summarizes the chosen task-condition pairs used as labels for conditional generation.

\begin{table}[h]
\centering
\caption{Selected task-conditions from the HCP task-fMRI dataset. One representative condition was chosen for each paradigm.}
\label{tab:task_conditions}
\begin{tabular}{ll}
\toprule
\textbf{Paradigm} & \textbf{Selected Condition} \\
\midrule
Working Memory  & 2-back places condition \\
Emotion         & Fear condition \\
Gambling        & Loss condition \\
Language        & Story condition \\
Relational      & Relation condition \\
Motor           & Right hand condition \\
Social          & Mental condition \\
\bottomrule
\end{tabular}
\end{table}

\subsection{Implementation Details}
\label{appx:implementation_details}

We fine-tuned the pretrained 3D VQ-GAN provided by \cite{kim2024adaptive} on individual HCP task-fMRI volumes and subsequently fixed it as an encoder for latent extraction. Diffusion models were trained with AdamW (learning rate $1\!\times\!10^{-4}$, weight decay $0.01$, $\beta_1{=}0.9$, $\beta_2{=}0.99$), batch size 16, and 400k steps. We used $T{=}1000$ diffusion steps with a linear noise schedule ($\beta_{\text{start}}{=}0.0015$, $\beta_{\text{end}}{=}0.0195$) and class dropout rate 0.05. 
For sample generation, we used EMA models with decay $0.9999$. 
All experiments were run on a single NVIDIA A100 GPU (40GB) with bfloat16 automatic mixed precision (AMP). 
Models were consistently evaluated at the 400k training checkpoint. Temporal frames were stacked along the channel dimension to enable joint spatio-temporal modeling across all models.

\begin{table}[h]
\centering
\caption{Model configurations for scalability study.}
\label{tab:scalability_config}
\begin{tabular}{lrrr}
\toprule
Model size & base\_ch & cond\_dim & num\_heads \\
\midrule
38.1 M   & 128 & 512  & 8  \\
85.4 M   & 192 & 768  & 12 \\
151.5 M  & 256 & 1024 & 16 \\
236.5 M  & 320 & 1280 & 20 \\
340.3 M  & 384 & 1536 & 24 \\
462.9 M  & 448 & 1792 & 28 \\
\bottomrule
\end{tabular}
\end{table}

\begin{table}[h]
\centering
\caption{Baseline MONAI 3D U-Net diffusion model configurations.}
\label{tab:baseline_config}
\begin{tabular}{lrr}
\toprule
Model size & num\_channels & num\_head\_channels \\
\midrule
41.3 M   & [64, 160, 224] & [0, 160, 224] \\
83.7 M   & [96, 224, 320] & [0, 224, 320] \\
141.0 M  & [128, 288, 416] & [0, 288, 416] \\
237.4 M  & [192, 352, 544] & [0, 352, 544] \\
475.8 M  & [256, 512, 768] & [0, 512, 768] \\
\bottomrule
\end{tabular}
\end{table}

\clearpage

\subsection{Evaluation Metrics}
\label{appx:eval_details}

\paragraph{GLM Activation Maps.}
We fit first-level generalized linear models (GLMs) at the subject level and second-level random-effects GLMs at the group level to obtain condition-specific statistical contrasts (\emph{z}-maps). For real data, GLMs were estimated on the held-out test set. For synthetic data, we generated 100 samples per condition and computed their group-level GLM maps analogously. Fidelity is measured as voxelwise Pearson correlation between real and synthetic group-level maps.

\paragraph{Representational Similarity Analysis (RSA).}
For each dataset (real, synthetic), we compute a representational dissimilarity matrix (RDM) $R \in \mathbb{R}^{K \times K}$, where $K$ is the number of task conditions. Each entry $R_{ij}$ is defined as the Pearson correlation between the group-level activation maps of task $i$ and $j$. We then vectorize the off-diagonal entries of the real and synthetic RDMs and compute their correlation. A high correlation indicates that the model has captured the higher-order representational geometry across tasks.

\paragraph{Condition Specificity (Top-1 Accuracy).}
For each generated condition, we compute voxelwise correlations with all real group-level maps and rank them. Top-1 accuracy is the fraction of cases where the correct condition achieves the highest correlation. This metric evaluates whether the generative model produces samples that are specific and distinguishable for their intended task condition.

\subsection{Additional Results}
\label{appx:additional_results}

\begin{figure}[h]  
    \centering
    \includegraphics[width=1.0\linewidth]{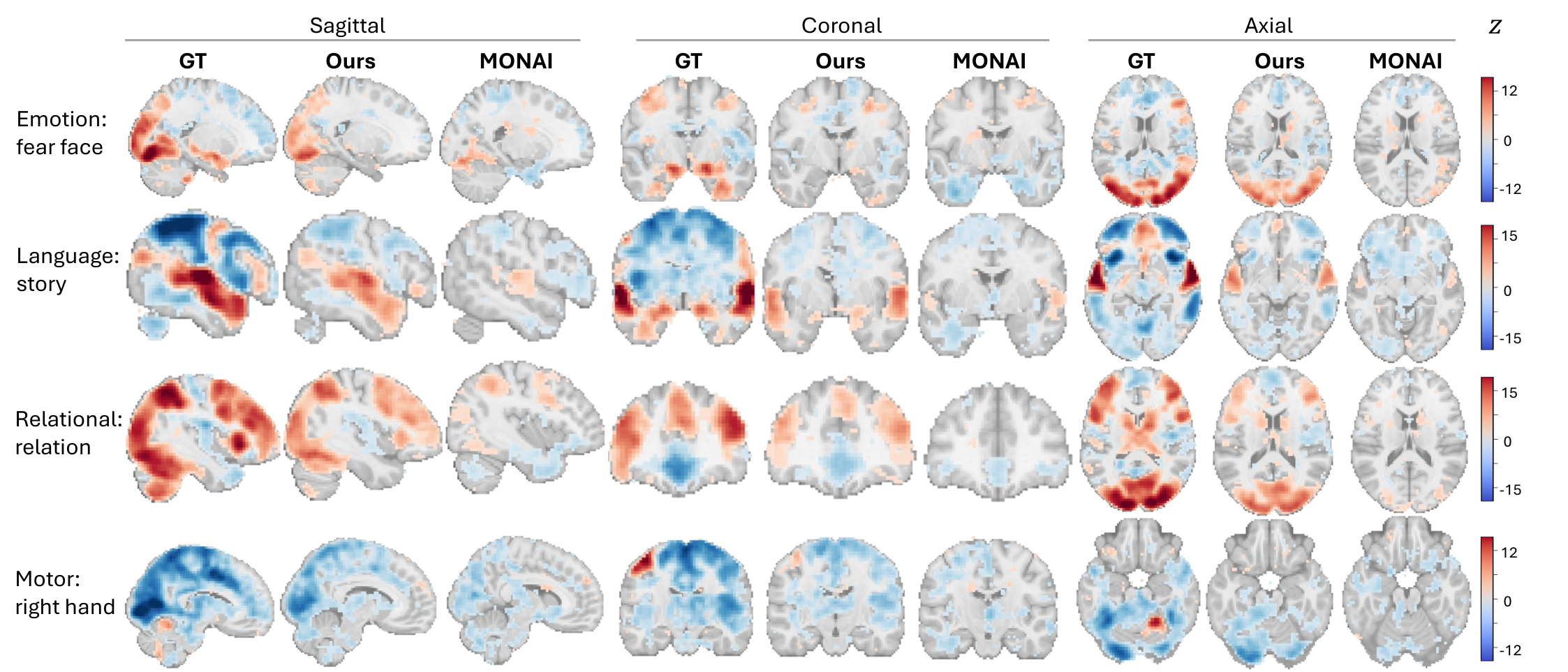}
    \caption{Additional visual results compared against MONAI~\cite{pinaya2023generative}.}
    \label{fig:supple_generation}
\end{figure}

Fig.~\ref{fig:supple_generation} shows comparisons of GLM Activation Maps for the other four tasks that could not be fully shown in Fig.~\ref{fig:generation}c due to page limit.

\clearpage
\subsection{ROI Time-Series Analysis}
\label{appx:roi_analysis}

To further evaluate whether our hybrid CNN–Transformer diffusion model captures task-specific temporal dynamics of brain activations, we examined ROI mean time-series in regions known to show significant activation or deactivation for each HCP task ~\cite{barch2013function}. ROI extraction was performed using the Harvard–Oxford cortical and subcortical atlas resampled to 3 mm resolution. For each task, we selected three representative ROIs spanning distinct functional systems.

\begin{figure}[h]  
    \centering
    \includegraphics[width=0.7\linewidth]{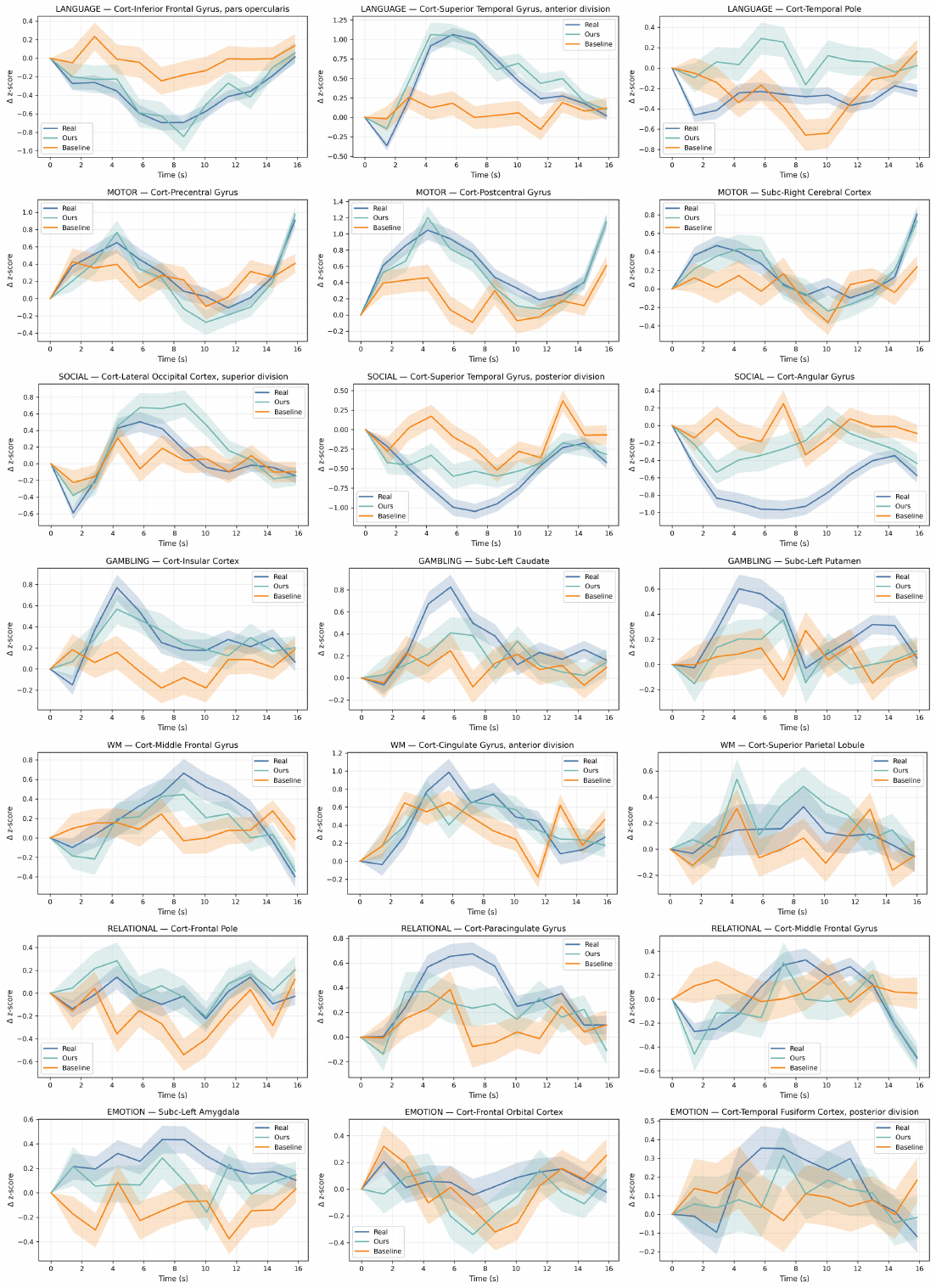}
    \caption{ROI mean time-series (mean ± SEM) for representative regions across seven HCP task contrasts (vs baseline). For each task, three ROIs were selected based on prior evidence of robust activation or deactivation~\cite{barch2013function}. Real test data (blue), synthetic data from our model (green), and the MONAI baseline (orange) are shown. Our model consistently better matches the temporal dynamics of real fMRI compared to the baseline.}
    \label{fig:roi_timeseries}
\end{figure}

Figure~\ref{fig:roi_timeseries} compares real fMRI data (test set not used in training) with synthetic data from our model and the MONAI baseline. Real and synthetic time-courses are plotted as mean ± SEM across samples. Across tasks, our model more faithfully reproduces the temporal profiles observed in real data, whereas the baseline often underestimates or distorts condition-specific responses. These results indicate that our approach not only preserves spatial activation patterns but also learns realistic temporal dynamics of task-evoked BOLD signals.

\end{document}